\documentclass[runningheads]{llncs}
\usepackage[T1]{fontenc}
\usepackage{graphicx,verbatim}
\usepackage{amssymb,amsfonts}
\usepackage{booktabs,multirow}
\usepackage{hyperref}
\usepackage{capt-of}
\newcommand{\XSolid}{\texttimes}
\usepackage{marvosym}
\usepackage{color}

\begin{document}
\titlerunning{Prenatal US Anomaly Classification and Localization}
\title{Prototype Memory-Guided Training-Free Anomaly Classification and Localization in Prenatal Ultrasound}
%

\author{
Huanwen Liang\inst{1,2}\thanks{Huanwen Liang and Yuhao Huang contributed equally.}
\and 
Yuhao Huang\inst{1,5\star} \and
Xiliang Zhu\inst{4} \and
Yuanji Zhang\inst{6} \and
Xuedong Deng\inst{7} \and
Xinru Gao\inst{8} \and
Guowei Tao\inst{9} \and
Yuhan Zhang\inst{1,2}$^{\href{zhangyuhan@szu.edu.cn}{\textrm{\Letter}}}$ \and
Dong Ni\inst{1,3,4}$^{\href{nidong@szu.edu.cn}{\textrm{\Letter}}}$
}

\authorrunning{H. Liang et al.}

\institute{
Medical Ultrasound Image Computing (MUSIC) Lab, Shenzhen University, Shenzhen, China
\and
School of Biomedical Engineering, Medical School, Shenzhen University, \\
Shenzhen, China
\and
School of Artificial Intelligence, Shenzhen University, Shenzhen, China
\and
School of Biomedical Engineering and Informatics, Nanjing Medical University, Nanjing, China
\and
Centre for Artificial Intelligence and Robotics, Hong Kong Institute of Science \& Innovation, Chinese Academy of Sciences, Hong Kong, China 
\and
Shenzhen Luohu People's Hospital (The Third Affiliated Hospital of Shenzhen University), Shenzhen, China.
\and
Center for Medical Ultrasound, The Affiliated Suzhou Hospital of Nanjing Medical University, Suzhou Municipal Hospital, Nanjing Medical University, Suzhou, China
\and
Northwest Women’s and Children’s Hospital, Xian, China
\and
Qilu Hospital of Shandong University, Jinan, China
}
\maketitle              
\begin{abstract}

Prenatal anomaly classification and localization is of critical importance for fetal health and pregnancy management.
Although ultrasound (US) is the primary modality for prenatal screening, accurate diagnosis remains challenging due to the low prevalence and high heterogeneity of anomalies. Existing deep learning methods for prenatal tasks rely on large-scale annotated datasets, which are difficult to obtain in practice.
Although few-shot learning alleviates data scarcity, it typically requires fine-tuning for new categories, limiting its practicality in resource-limited clinical settings.
To address these challenges, we propose a training-free framework for multi-class prenatal US anomaly classification and localization that operates with only a few reference images per class, representing the first exploration of this setting.
Our framework comprises three key components:
(1) a memory bank with multi-granular prototypes that explicitly models both class-level semantics and anomaly characteristics;
(2) a prototype-driven soft merging mechanism that aggregates discriminative features to detect the anomaly region; and 
(3) a class-aware refinement strategy that leverages prototype consistency to improve category prediction.
Extensively validated on a multi-center prenatal US dataset containing 1,149 cases, with a total of 2,357 images and 9 categories, our proposed method outperforms the competitors.


\end{abstract}
\section{Introduction}

Prenatal anomalies are functional or structural abnormalities that occur during fetal growth~\cite{laurichesse2017congenital,pechriggl2022embryology}. These anomalies can seriously affect fetal survival and postnatal quality of life~\cite{xie2025global}. Therefore, accurate diagnosis is critical for pregnancy management. 
Ultrasound (US) is widely used for prenatal anomaly screening and diagnosis because it is non-invasive, radiation-free, and enables real-time imaging~\cite{salomon2022isuog,bilardo2023isuog}. 
However, the low incidence and high categorical diversity of prenatal anomalies lead to heterogeneous US appearances, making accurate identification difficult even for experienced sonographers.
Consequently, prenatal anomaly screening becomes highly operator-dependent and susceptible to substantial inter-operator variability, underscoring the clinical value of automated methods for assisting screening and reducing diagnostic variability.

\textbf{Learning-based method} in prenatal US focuses on various tasks including plane detection~\cite{yang2021searching,dou2025standard}, biomarker measurement~\cite{zhang2025deep}, and disease diagnosis~\cite{huang2025uncertainty,zhang2026artificial}.
Zhang et al.~\cite{zhang2025comparative} proposed a deep learning-based method to identify the standard fetal midsagittal plane in 3D US volumes for crown-rump length measurement. 
Lin et al.~\cite{lin2022use} developed a YOLOv3-based system to detect fetal brain standard planes and to classify abnormal images. 
Liang et al.~\cite{liang2025medical} introduced medical knowledge to enhance anomaly recognition and achieved prenatal abdominal anomaly classification without relying on standard planes.
These studies have shown strong performance in prenatal US tasks. However, most methods rely on large amounts of annotated data. Since prenatal anomalies have a low incidence and data access is constrained by patient privacy, constructing large-scale annotated datasets is costly. This hinders the further development of these methods.

\begin{figure}[!t]
\includegraphics[width=\textwidth]{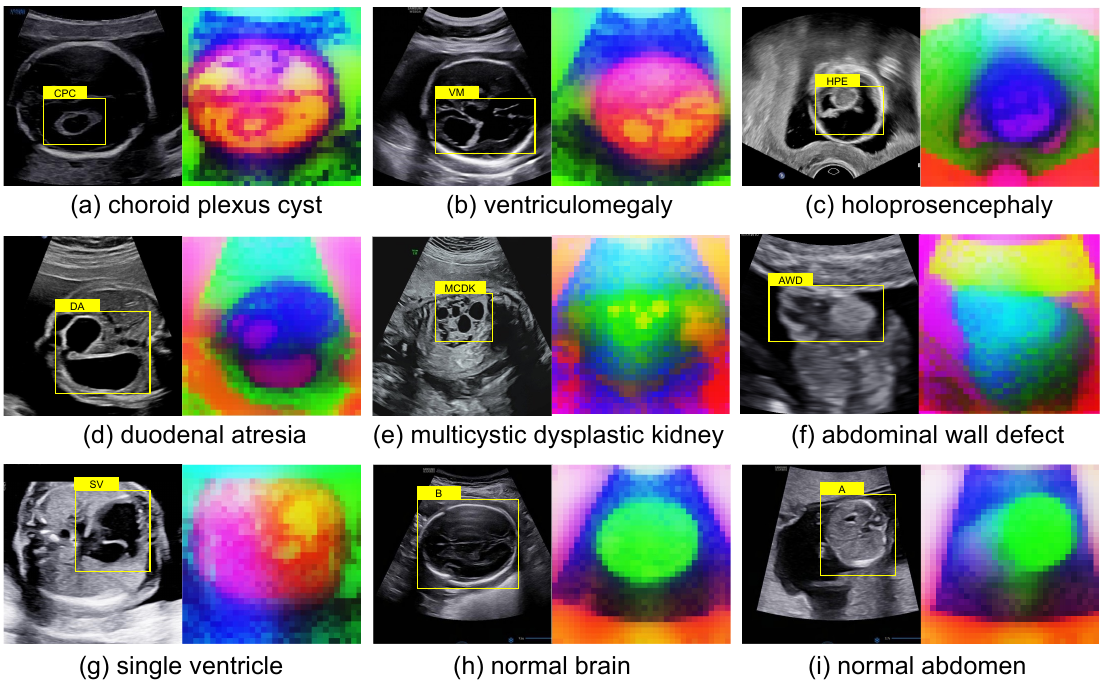}
\centering
\caption{Visualization of DINO features. Each image (right) displays the first three
principal components of the feature representation, mapped to RGB color channels.
yellow boxes indicate region of interest.} \label{fig1}
\end{figure}

\textbf{Few-shot learning}, e.g., few-shot object detection (FSOD), is considered an effective solution to alleviate data scarcity~\cite{defrcn,digeo,smile,lu2023breaking}.
SMILe~\cite{smile} introduces submodular mutual information as the learning objective to encourage well-separated feature clusters between base and novel classes, thereby reducing class confusion and catastrophic forgetting in FSOD.
Zhu et al.~\cite{trr} propose TRR-CCM, an US FSOD method that leverages circular channel Mamba and topological relationship reasoning to improve detection and localization accuracy.
However, existing methods rely on large-scale annotated data for pretraining from the same modality. They also require fine-tuning on novel categories, which limits their practicality in resource-limited clinical settings.

\textbf{Foundation vision models (FVMs)}, such as Contrastive Language-Image Pre-training (CLIP)~\cite{clip} and Self-Distillation with No Labels (DINO)~\cite{dino1,dino2,dino3}, have recently achieved strong representation learning ability through pretraining on large-scale datasets. 
Recent studies have explored applying these foundation models to training-free medical image segmentation~\cite{maup,liu2025synpo,yang2025medsamix}, showing their potential in low-annotation or annotation-free settings. 
As shown in Fig.~\ref{fig1}, although DINO is not trained on US data, it can still capture fetal structural features, demonstrating strong fine-grained representation ability.

In this study, we propose a training-free framework for prenatal anomaly classification and localization based on vision foundation encoder. Our framework does not require fine-tuning and does not rely on large amounts of annotated data, providing a practical solution for prenatal anomaly classification and localization.
Our contributions are threefold.
(1) To the best of our knowledge, this is the first study to perform multi-disease prenatal anomaly classification and localization in a training-free setting without relying on extensive annotated data.
(2) We develop a unified framework that jointly incorporates memory bank construction, anomaly feature aggregation, and class-aware refinement to facilitate accurate and efficient localization of abnormal regions.
(3) We validated our approach on a multi-center prenatal anomaly US dataset covering 9 categories, with 1,149 cases and 2,357 images. 
Experimental results demonstrated the effectiveness of our method.

\begin{figure}[!t]
\includegraphics[width=\textwidth]{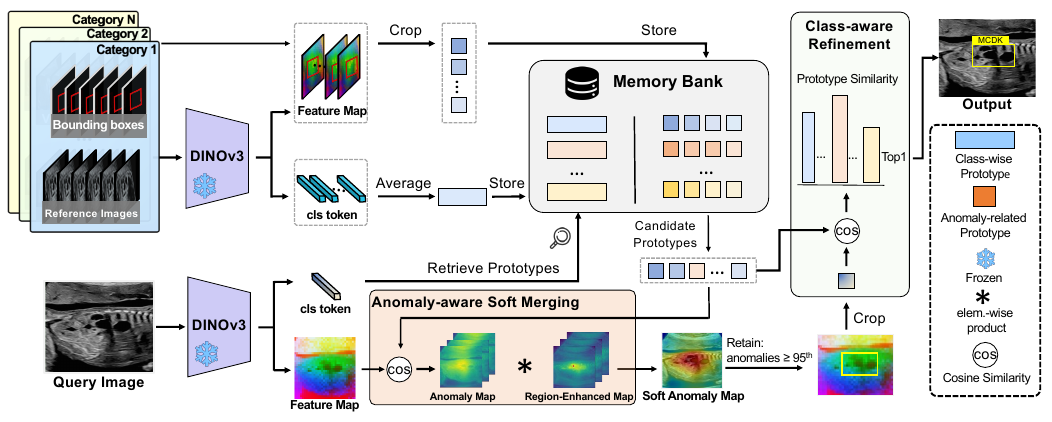}
\caption{Overview of our proposed framework.} \label{framework}
\end{figure}

\section{Methodology}
Fig.~\ref{framework} illustrates the overall framework of the proposed training-free method, which relies on only a few annotated reference images for accurate anomaly classification and localization.
Our model consists of three stages.
In \textbf{Stage 1}, a memory bank was constructed to store class-wise and anomaly-related prototype representations.
Then, \textbf{Stage 2} introduced an anomaly feature aggregation strategy that selects candidate anomaly-related prototypes based on class-wise feature similarity and computes anomaly scores through anomaly-feature matching with anomaly-aware soft merging.
Last, a class-aware refinement strategy was equipped to improve the anomaly category discrimination in \textbf{Stage 3}.

\subsection{Multi-Granular Prototype Memory Bank Construction}
A good vision foundation encoder $E$ plays a vital role in extracting features for memory bank construction.
Most recently, DINOv3 has shown powerful performance to provide visual features for multiple medical imaging tasks~\cite{liu2025does}.
Inspired by this study, we employ DINOv3~\cite{dino3} as the encoder $E$ in our work. 

Given a set of reference images $\{I_i\}_{i=1}^{N}$ with anomaly bounding box annotations 
$\{B_i = \{(x_1, y_1), (x_2, y_2)\}\}_{i=1}^{N}$, where $N = C \times K$, $C$ denotes the number of categories, 
$K$ denotes the number of reference images per category, $(x_1, y_1)$ and $(x_2, y_2)$ denote the 
top-left and bottom-right coordinates of the bounding box, respectively.
DINOv3 first partitions $I_i$ into patches and processes them through $L$ transformer layers, producing patch embeddings $z^l, l\in L$ and a corresponding $[\mathrm{CLS}]^l$ token at each layer.
\begin{equation}
E = \{layer^1, layer^2, \cdots, layer^L\},
\end{equation}
\begin{equation}
\{[\mathrm{CLS}]^l, z^l \} = layer^l(z^{l-1}),z^0 = I_i.
\end{equation}

\textbf{Class-wise Prototype Construction for Coarse-grained Recognition.}
For each reference image $I_i$, we pass it through the encoder $E$ and extract the $[\mathrm{CLS}]^L$ token from the last layer.
After $\ell_2$ normalization, the token is stored as a global class feature $g_i \in \mathbb{R}^{D}$, where $D$ denotes the feature dimension.
Then, for each category $c\in C$, class features from reference images belonging to the same category are averaged to form a class-wise prototype:
\begin{equation}
\mathbf{p}_c = \frac{1}{K} \sum_{i \in \mathcal{I}_c}g_i,
\end{equation}
where $\mathcal{I}_c$ denotes the index set of category $c$.
These class-wise prototypes are stored in the memory bank as class-wise representations.

\textbf{Anomaly-Related Prototype Construction for Fine-grained Identification.}
For each reference image $I_i$ with anomaly bounding box annotation $B_i$, 
we pass $I_i$ through the encoder $E$ and extract $\ell_2$-normalized patch embeddings $z^L$ from the last layer as dense feature map 
$\mathbf{F_i} \in \mathbb{R}^{d \times h \times w}$.
Due to the low resolution of the original feature maps, a high-resolution upsampling method, JAFAR~\cite{jafar}, is adopted to enhance spatial details to obtain high-resolution feature map $\mathbf{F_{hr}}_i$.
For each image, we aggregate features within the annotated bounding box to form an anomaly-related prototype:
\begin{equation}
\mathbf{q}_i = \frac{1}{|B_i|} \sum_{(x,y)\in B_i} \mathbf{F_{hr}}_i(x,y).
\end{equation}
These anomaly-related prototypes capture localized anomaly features and complement the class-wise prototypes stored in the memory bank, thereby forming a multi-granular prototype representation and improving anomaly localization.

\subsection{Prototype-Guided Anomaly Feature Aggregation}
During inference, given a query image $I_q$, its class feature is extracted using the encoder $E$ and matched with the class-wise prototypes to select top-$M$ candidate categories $\mathbf{p}_c$.
The corresponding anomaly-related prototypes are then retrieved from the memory bank for anomaly aggregation.

\textbf{Anomaly-Feature Matching.}
First, we retrieve candidate anomaly-related prototypes $\{\mathbf{q}_{c,k}\}$.
Then, we extract dense feature map $\mathbf{F}_q$ from the query image and upsample to obtain high-resolution feature map $\mathbf{F}_{hr}$.
We compute the cosine similarity between each candidate prototype and $\mathbf{F}_{hr}$ (see Equ.~\ref{equ: cos_anomaly}), resulting in one anomaly map per retrieved prototype.
\begin{equation}
\label{equ: cos_anomaly}
\mathbf{A}_{c,k} = \cos\left(\mathbf{q}_{c,k}, \mathbf{F}_{hr}\right).
\end{equation}

\textbf{Anomaly-Aware Soft Merging.}
In US images, distinct anatomical structures can exhibit visually similar feature patterns, resulting in feature-level ambiguity that introduces noise and leads to false detections when features are matched independently.
Meanwhile, we noticed that a local patch feature is often highly similar to its neighboring patches, which can be leveraged to strengthen region-level attention.
Hence, for each anomaly map, our anomaly-aware soft merging strategy first selects the top-$S$ high-similarity locations and computes their centroid $\mathbf{f}_{c,k}$.
Then, the cosine similarity between the centroid feature $\mathbf{f}_{c,k}$ and $\mathbf{F}_{hr}$ was evaluated to obtain the region-enhanced map:
\begin{equation}
\mathbf{R}_{c,k} = \cos\left(\mathbf{f}_{c,k}, \mathbf{F}_{hr}\right),
\end{equation}
which can be used to softly merge the $K$ anomaly maps:
\begin{equation}
\mathbf{A}_c = \frac{1}{K}\sum_{k=1}^{K} \mathbf{R}_{c, k}\,\mathbf{A}_{c,k}
\end{equation}

\textbf{Anomaly Localization.}
Patches with anomaly values above the 95th percentile threshold in $\mathbf{A}_c$ are retained to generate a binary mask.
Then, the minimum enclosing rectangle is computed as the final anomaly regions $\Omega$.

\subsection{Class-Aware Refinement with Enhanced Memory Alignment}
Based on the anomaly regions obtained in the previous step, this stage further refines their categories via a class-aware refinement strategy with the constructed memory bank.
For each predicted anomaly region, we extract its region features from $\mathbf{F}_{hr}$ and compute an average feature embedding:
\begin{equation}
\mathbf{e}_i = \frac{1}{|\Omega_i|} \sum_{(x,y)\in \Omega_i} \mathbf{F}_{hr}(x,y),
\end{equation}
where $\Omega_i$ denotes the bounding box of the $i$-th anomaly region.
We then compute the cosine similarity between $\mathbf{e}_i$ and the retrieved anomaly-related prototypes, and select the class with the highest similarity as the predicted label, with the similarity value serving as the confidence score.

\begin{table}[!t]
\centering
\caption{Quantitative comparison under the 32-shot setting across Splits 1-5 in terms of mAP@0.5 ($ p < 0.05$).
\textbf{Bold} indicates the best performance. 
All results are reported as (mean(std)) (\%). TF: Training-free.}
\label{tab:comparison}
\renewcommand{\arraystretch}{1}
\resizebox{\linewidth}{!}{
\begin{tabular}{l|l|c|c|c|c|c|c|c|c|c|c}
\hline
\multirow{2}{*}{} & \multirow{2}{*}{} & CPC & VM & HPE & SV & DA & MCDK & AWD & B & A & mAP \\
\hline
\multirow{8}{*}{FSOD} & \multirow{2}{*}{DeFRCN}     & \textbf{52.60} & 39.23 & 55.63 & 50.18 & 52.97 & 35.18 & \textbf{42.34} & 36.16 & 54.90 & 46.58 \\
~ & ~ & (1.78) & (2.94) & (1.29) & (6.19) & (4.91) & (9.38) & (2.51) & (2.59) & (2.15) & (2.44) \\
\cmidrule(lr){2-12}

& \multirow{2}{*}{DiGeo}      & 48.81 & 38.77 & 40.00 & 69.03 & 54.47 & 34.09 & 28.50 & 38.52 & 57.12 & 45.48 \\
~ & ~ & (4.75) & (7.46) & (4.42) & (10.21) & (6.73) & (10.39) & (6.73) & (3.17) & (0.82) & (4.46) \\
\cmidrule(lr){2-12}

& \multirow{2}{*}{SMILe-FSOD} & 50.5 & 44.11 & 41.40 & 62.89 & 53.91 & 39.22 & 32.18 & 39.45 & 54.64 & 46.48 \\
~ & ~ & (1.29) & (2.13) & (1.19) & (7.21) & (5.62) & (7.03) & (4.23) & (1.94) & (2.36) & (2.58) \\
\cmidrule(lr){2-12}

& \multirow{2}{*}{TRR-CCM}    & 51.74 & 36.79 & 53.12 & 71.13 & 53.8 & 38.36 & 40.75 & 43.18 & 54.52 & 49.26 \\
~ & ~ & (4.20) & (4.56) & (4.49) & (4.89) & (4.42) & (6.70) & (1.20) & (2.62) & (3.56) & (2.79) \\

\hline
\multirow{6}{*}{TF} & \multirow{2}{*}{ProtoSAM}   & 22.57 & 38.13 & 26.66 & 28.85 & 55.96 & 55.8 & 10.23 & 64.65 & 65.71 & 40.95 \\
~ & ~ & (3.39) & (3.33) & (3.71) & (5.97) & (6.44) & (6.27) & (7.78) & (7.67) & (5.62) & (2.86) \\
\cmidrule(lr){2-12}

& \multirow{2}{*}{MAUP}       & 25.46 & 37.61 & 15.89 & 38.17 & 35.02 & 39.72 & 17.01 & \textbf{66.93} & 55.22 & 36.78 \\
~ & ~ & (4.32) & (7.01) & (7.21) & (8.86) & (4.22) & (8.58) & (5.43) & (6.95) & (3.47) & (2.34) \\
\cmidrule(lr){2-12}

& \multirow{2}{*}{Ours}       & 50.66 & \textbf{71.00} & \textbf{58.19} & \textbf{71.29} & \textbf{76.77} & \textbf{64.67} & 36.17 & 46.39 & \textbf{70.88} & \textbf{60.67} \\
~ & ~ & (6.35) & (3.56) & (5.12) & (5.59) & (1.44) & (2.16) & (2.43) & (2.70) & (4.91) & (1.82) \\
\hline
\end{tabular}
}
\end{table}

\section{Experiment Results}
\textbf{Materials and Implementation Details.}
With local Institutional Review Board approval (\textit{No.} K-2023-067-H01), we collected a multi-center prenatal US dataset comprising 1,149 cases and 2,357 images, encompassing 9 categories spanning 3 anatomical regions (brain, heart, and abdomen).
All images were annotated by sonographers with classification labels and bounding boxes: the anomaly class is annotated with the anomaly region, and the normal class is annotated with the region of interest.
Specifically, the fetal brain images contain four types, including choroid plexus cyst (CPC, 122 cases / 226 images), ventriculomegaly (VM, 164 cases / 312 images), holoprosencephaly (HPE, 104 cases / 238 images), and normal brain (B, 183 cases / 329 images). 
The heart-related category contain single ventricle (SV, 29 cases / 197 images).
The abdomen-related categories contain duodenal atresia (DA, 178 cases / 255 images), multicystic dysplastic kidney (MCDK, 77 cases / 261 images), abdominal wall defect (AWD, 51 cases / 181 images), and normal abdomen (A, 241 cases / 358 images).
The dataset was split at the case level by sampling cases without replacement. Once the accumulated number of images exceeded 32, 32 images were randomly selected to form a fold, and the corresponding cases were removed from the pool. The process was repeated to construct five reference folds (Splits 1-5), each containing 32 images per category (1,440 images in total), while the remaining cases were used for testing (917 images in total).
Within each reference set, we randomly selected 4, 8, 16, or all 32 images per category to construct the 4-shot, 8-shot, 16-shot, and 32-shot settings, respectively.

Our approach was implemented in PyTorch using an NVIDIA RTX 4090 GPU (24GB).
Images were resized to $512 \times 512$.
We employed a pretrained DINOv3-Small backbone, producing $64 \times 64$ feature map.
The JAFAR upsampling method was applied to upsample features to $256 \times 256$ without additional pre-training. For hyperparameters, we set $M=2$ and $S=16$.
Performance was evaluated using mAP@0.5 with TorchMetrics. Codes can be found in~\href{https://github.com/LL-AC/TFF}{GitHub}.

\begin{table}[!t]
  \centering
  \begin{minipage}{0.4\textwidth}    
    \centering
    \captionof{table}{Ablation study results.}
    \renewcommand{\arraystretch}{1}
    \label{tab:ablation}
    \begin{tabular}{ccc|c}
    \hline
    HR & ASM & CR & mAP \\
    \hline
    \XSolid & \XSolid & \XSolid & 45.18$\pm$1.89 \\
    \checkmark & \XSolid & \XSolid & 47.90$\pm$1.56 \\
    \checkmark & \checkmark & \XSolid & 55.71$\pm$1.39 \\
    \checkmark & \XSolid & \checkmark & 53.37$\pm$1.06 \\
    \checkmark & \checkmark & \checkmark & 60.67$\pm$1.82 \\
    \hline
    \end{tabular}
    
    \centering
    \includegraphics[width=\linewidth]{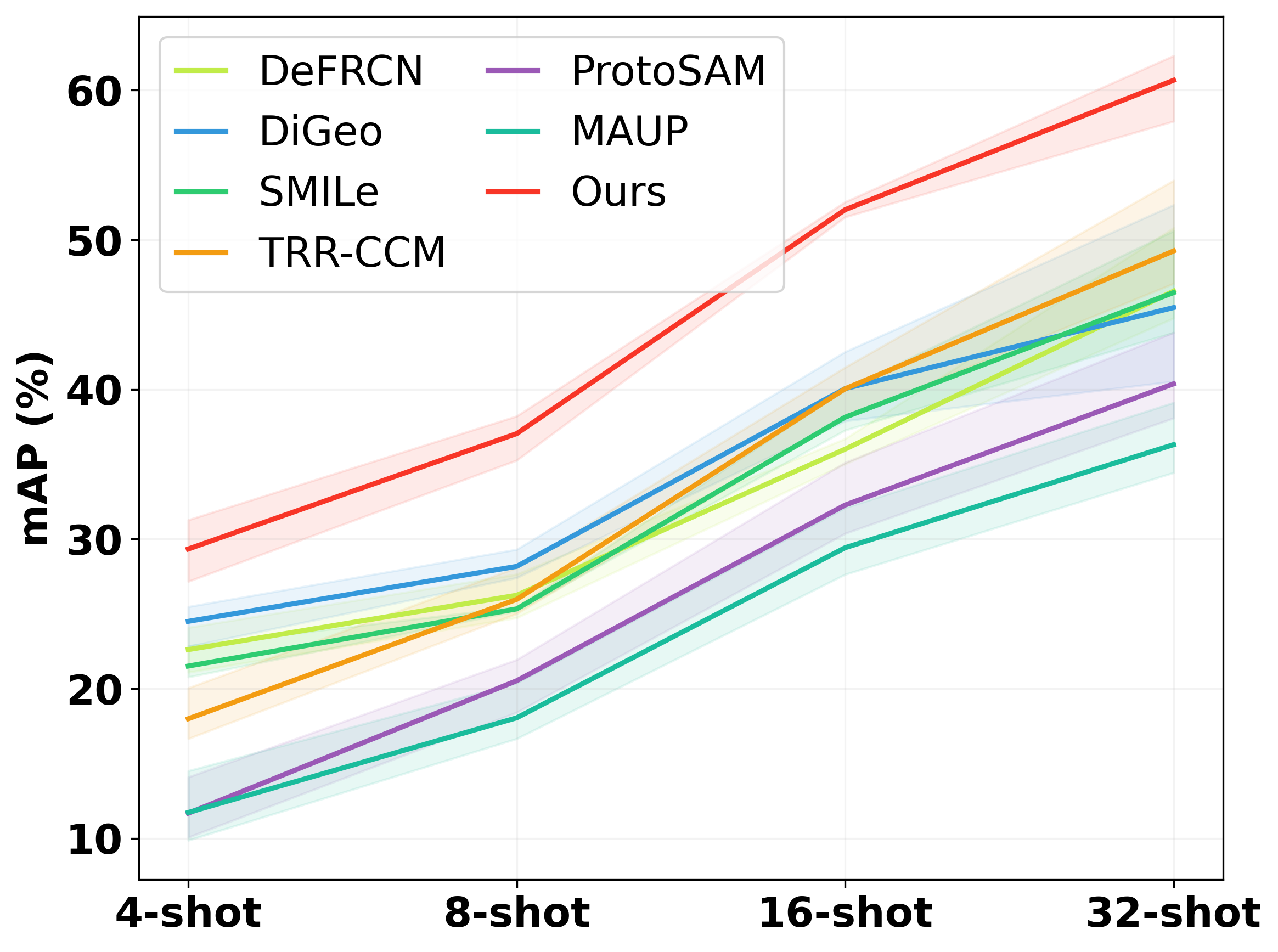} 
    \captionof{figure}{Performance visualization.} 
    \label{fig:few}
  \end{minipage}
  \begin{minipage}{0.55\textwidth}
    \centering
    \includegraphics[width=\linewidth]{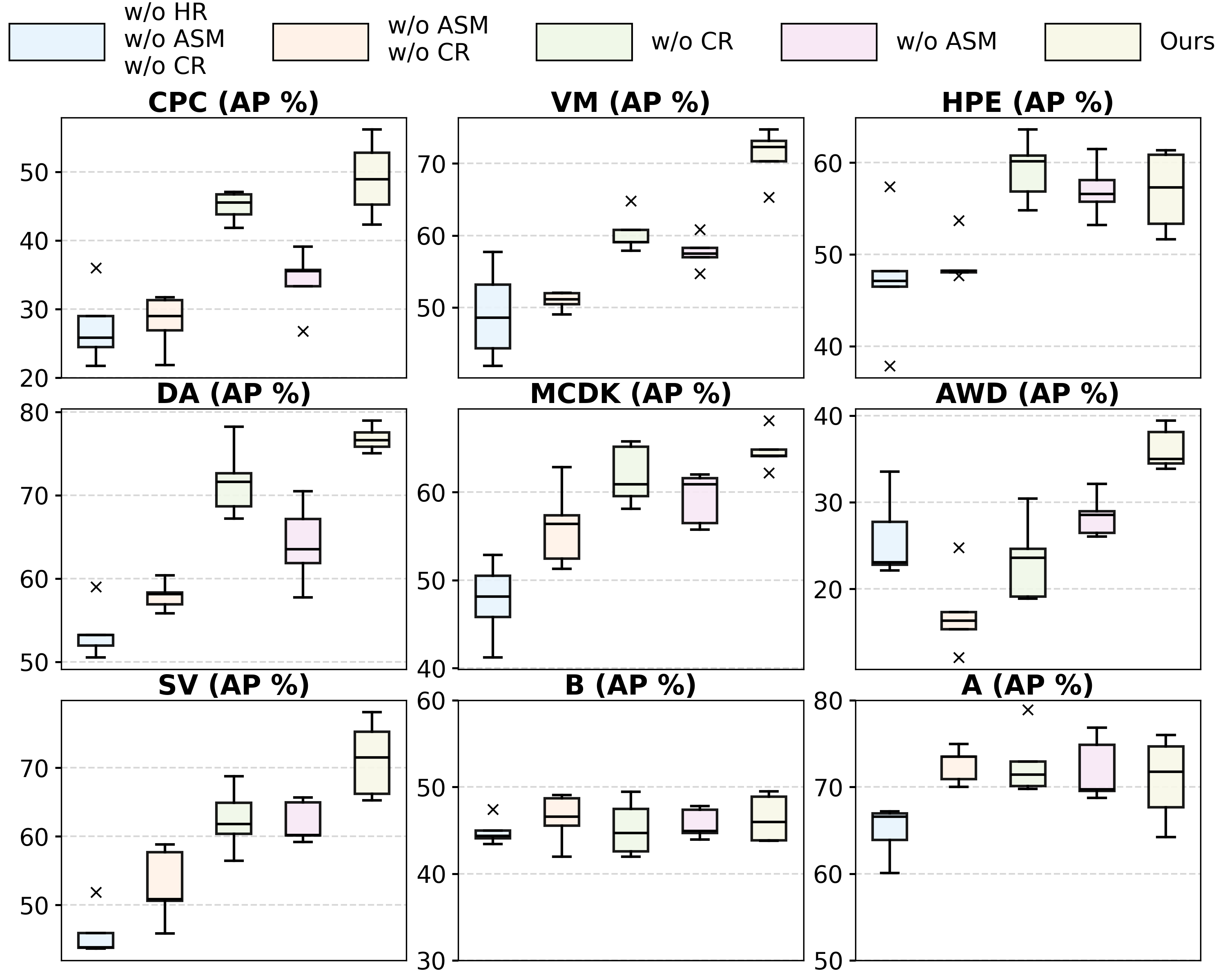} 
    \captionof{figure}{Ablation study on different categories.}
    \label{fig:visual}
  \end{minipage}
\end{table}

\textbf{Quantitative Analysis.}
We compare our approach with official implementations of FSOD methods, including DeFRCN~\cite{defrcn}, DiGeo~\cite{digeo}, SMILe-FSOD~\cite{smile}, and TRR-CCM~\cite{trr}, which are fine-tuned on our reference sets, as well as training-free methods, including ProtoSAM~\cite{protosam} and MAUP~\cite{maup}. Table~\ref{tab:comparison} shows the results.
Our method outperforms the comparative methods across most categories.
Compared with the second-best method, our approach achieves an 11.41\% higher mAP.
Fig.~\ref{fig:few} shows the mAP performance with error estimation across splits. As the number of reference images increases, our method exhibits steady performance gains, demonstrating strong scalability and outperforming all compared methods overall.
Table~\ref{tab:ablation} shows the ablation results.
High-resolution (HR) upsampling improves mAP by 2.72\%. Anomaly-aware soft merging (ASM) improves mAP by 7.81\%. Class-aware refinement (CR) improves mAP by 5.47\%. Using all components improves mAP by 15.49\%.
Fig.~\ref{fig:visual} shows the ablation results on different categories.
It can be observed that each proposed strategy provides systematic performance improvements across all categories. 

\begin{figure}[!t]
\includegraphics[width=0.93\textwidth]{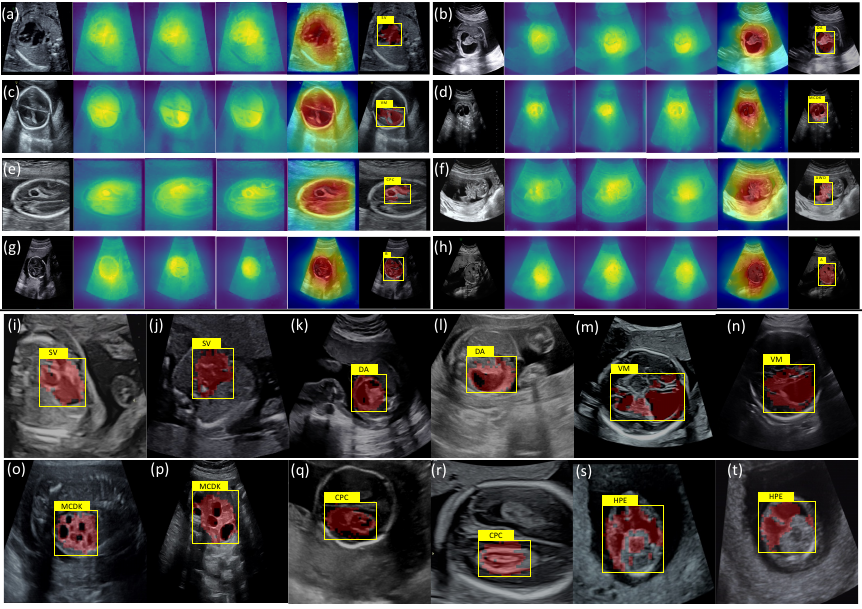}
\centering
\caption{Visualization of typical cases. (a)-(h) show representative examples. From left to right: the query image, similarity maps of three prototypes, the anomaly map, and the predicted anomaly region with category label. (i)-(t) shows additional cases.
}
\label{visual}
\end{figure}

\textbf{Qualitative Analysis.}
We visualize several typical cases in Fig.~\ref{visual}.
For each query image (a)-(h), we show three candidate prototypes.
We observe that, after aggregating the similarity maps from these candidate prototypes, the resulting anomaly map is well concentrated on the fetal region.
The mask and bounding box obtained using the 95th-percentile threshold accurately localize the anomaly region.
These results indicate that, even without training and using only a few reference images, our approach can accurately localize anomaly regions and correctly classify their categories, highlighting its effectiveness.

\section{Conclusion}
In this paper, we proposed a novel training-free framework for multi-class prenatal anomaly classification and localization.
Our method requires only a few reference images, addressing the scarcity of abnormal data and eliminating the need for retraining.
The proposed framework consists of three stages: (1) memory bank construction to store class-level and anomaly-related representations from reference images. (2) anomaly feature aggregation to localize abnormal regions via anomaly-aware soft merging. (3) class-aware refinement to improve anomaly category discrimination.
We evaluated our method on a multi-center dataset using only a few reference images per class.
The results demonstrated the effectiveness of our approach.
In future work, we will extend our approach to additional anomaly types and perform comprehensive clinical validation.

\begin{credits}
\subsubsection{\ackname} This work is supported by the Frontier Technology Development Program of Jiangsu Province (No. BF2024078), National Natural Science Foundation of China (Nos. 12326619, 62572324), Science and Technology Planning Project of Guangdong Province (No. 2023A0505020002).

\subsubsection{\discintname} The authors have no competing interests to declare that are relevant to the content of this article.

\end{credits}

%
%
%
\bibliographystyle{splncs04}
\bibliography{Paper}
%




\end{document}